\documentclass[11pt]{article} 
\usepackage{url}
\usepackage{smile}
\usepackage{graphicx} 
\usepackage{algorithm}
\usepackage{algorithmic}
\usepackage{todonotes}
\usepackage{epstopdf}
\usepackage{wrapfig}
\usepackage[colorlinks, linkcolor=blue, anchorcolor=blue, citecolor=blue]{hyperref}
\usepackage[margin=1in]{geometry}
\usepackage[normalem]{ulem}
\usepackage[export]{adjustbox}
\usepackage{mathtools, cuted}
\usepackage{natbib}
\usepackage{enumerate}
\usepackage{microtype}
\usepackage{graphicx}
\usepackage{subcaption}
\usepackage{booktabs} 
\usepackage{smile}
\usepackage{wrapfig}

\usepackage{kpfonts}
\DeclareMathAlphabet{\mathsf}{OT1}{cmss}{m}{n}
\SetMathAlphabet{\mathsf}{bold}{OT1}{cmss}{bx}{n}

\providecommand{\norm}[1]{\|#1\|}

\begin{document}

\title{\huge On Scalable and Efficient Computation of Large Scale Optimal Transport}

\author{Yujia Xie, Minshuo Chen, Haoming Jiang, Tuo Zhao, Hongyuan Zha \thanks{Authors are affiliated with Georgia Institute of Technology. Emails: \{\tt xieyujia, mchen393, jianghm, tourzhao\}@gatech.edu, zha@cc.gatech.edu}}
\date{}

\maketitle



\begin{abstract}

Optimal Transport (OT) naturally arises in many machine learning applications, yet the heavy computational burden limits its wide-spread uses. To address the scalability issue, we propose an implicit generative learning-based framework called SPOT (Scalable Push-forward of Optimal Transport). Specifically, we approximate the optimal transport plan by a pushforward of a reference distribution, and cast the optimal transport problem into a minimax problem. We then can solve OT problems efficiently using primal dual stochastic gradient-type algorithms. We also show that we can recover the density of the optimal transport plan using neural ordinary differential equations. Numerical experiments on both synthetic and real datasets illustrate that SPOT is robust and has favorable convergence behavior. SPOT also allows us to efficiently sample from the optimal transport plan, which benefits downstream applications such as domain adaptation.

\end{abstract}

\section{Introduction}

The Optimal Transport (OT) problem naturally arises in a variety of machine learning applications, where we need to handle data from multiple sources. One example is domain adaptation: We collect multiple datasets from different domains, and we need to learn a model from a source dataset, which can be further adapted to target datasets \citep{ganin2014unsupervised,courty2017optimal,damodaran2018deepjdot}. Another example is resource allocation: We want to assign a set of assets (one data source) to a set of receivers (another data source) so that an optimal economic benefit is achieved \citep{santambrogio2010models,galichon2017survey}. Recent literature has shown that both aforementioned applications can be formulated as optimal transport problems.

The optimal transport problem has a long history, and its earliest literature dates back to \citet{monge1781memoire}. Since then, it has attracted increasing attention and been widely studied in multiple communities such as applied mathematics, probability, economy and geography \citep{villani2008optimal, carlier2012optimal,gross2016recent}. Specifically, we consider two sets of $d$-dimensional data, which are generated from two different distributions denoted by $X\sim\mu$ and $Y\sim\nu$.\footnote{The optimal transport can also handle more than two distributions. See Section \ref{sec:framework} for more details.} We aim to find an optimal joint distribution $\gamma$ of $X$ and $Y$, which minimizes the expectation on some cost function $c$, i.e., 
\begin{align}
\label{eq:intro}
\gamma^*= \argmin_{\gamma \in \Pi(\mu,\nu)} \mathbb{E}_{(X,Y)\sim \gamma} [c(X,Y)],
\end{align}
The constraint $\gamma \in \Pi(\mu,\nu)$ requires the marginal distribution of $X$ and $Y$ in $\gamma$ to be identical to $\mu$ and $\nu$, respectively. Existing literature often refers to the optimal expected cost $\mathcal{W}^*(\mu,\nu)=\mathbb{E}_{(X,Y)\sim \gamma^*} [c(X,Y)]$ as \textit{Wasserstein distance}, and $\gamma^*$ as the \textit{optimal transport plan}. For domain adaptation, the function $c$ measures the discrepancy between $X$ and $Y$, and the optimal transport plan $\gamma^*$ essentially reveals the transfer of the knowledge from source $X$ to target $Y$. For resource allocation, the function $c$ is the cost of assigning resource $X$ to receiver $Y$, and the optimal transport plan $\gamma^*$ essentially yields the optimal assignment.

Since \eqref{eq:intro} is an optimization problem over the space of distributions, the problem is infinite dimensional and generally intractable when $\mu$ and $\nu$ are continuous distributions.
Therefore, existing literature has resorted to finite dimensional approximations. For example, \citet{cuturi2013sinkhorn} propose to discretize the support using a refined grid, and cast \eqref{eq:intro} into a finite dimensional linear programming problem. However, for complex distributions in high dimensions (e.g., images in domain adaptation), the grid size often needs to be exponentially large (e.g., exponential in dimension) to ensure a small approximation error (due to discretization). Under such a regime, conventional linear programming algorithms do not scale well, e.g., the interior point method in conjunction with the Newton's method takes ${\cO}(n^3 \log n)$ time, where $n$ is the grid size. To ease such a scalability issue, \citet{cuturi2013sinkhorn} propose an entropy regularization-based Sinkhorn algorithm, which requires the computational cost of ${\cO}(n^2)$, but still fail to scale to large problems. 

While there exist several scalable stochastic algorithms for computing Wasserstein distance for continuous distributions $\mu$ and $\nu$ \citep{genevay2016stochastic, seguy2017large, yang2018scalable}, they cannot compute the optimal transport plan $\gamma^*$ (see Section \ref{sec:discussion} for more discussion), which is crucial in the aforementioned applications. 



To address the scalability and efficiency issues, we propose a new implicit generative learning-based framework for solving optimal transport problems. Specifically, we approximate $\gamma^*$ by a generative model, which maps from some latent variable $Z$ to $(X,Y)$. For simplicity, we denote
\begin{align}\label{nn-generator}
\left[
\begin{array}{c}
X\\
\hline
Y
\end{array}
\right]
=G(Z)=\left[
\begin{array}{c}
G_X(Z)\\
\hline
G_Y(Z)
\end{array}
\right]\quad\textrm{with}\quad Z\sim\rho,
\end{align}
where $\rho$ is some simple latent distribution and $G$ is some operator, e.g., deep neural network or neural ordinary differential equation (ODE).
Accordingly, instead of directly estimating the probability density of $\gamma^*$, we estimate the mapping $G$ between $Z$ and $(X,Y)$ by solving
\begin{align}\label{generative-learning}
G^*=~&\argmin_G \quad~~ \EE_{Z\sim\rho}[c(G_X(Z),G_{Y}(Z))], \quad \textrm{subject~to}\quad G_{X}(Z)\sim\mu,~G_{Y}(Z)\sim\nu
\end{align}
We then cast \eqref{generative-learning} into a minimax optimization problem using the Lagrangian multiplier method. As the constraints in \eqref{generative-learning} are over the space of continuous distributions, the Lagrangian multiplier is actually infinite dimensional. Thus, we propose to approximate the Lagrangian multiplier by deep neural networks, which eventually delivers a finite dimensional generative learning problem.

Our proposed framework has three major benefits: (1) Our formulated minimax optimization problem can be efficiently solved by primal dual stochastic gradient-type algorithms. Many empirical studies have corroborated that these algorithms can easily scale to very large minimax problems in machine learning \citep{brock2018large}; (2) Our framework can take advantage of recent advances in deep learning. Many empirical evidences have suggested that deep neural networks can effectively adapt to data with intrinsic low dimensional structures \citep{zhang2016understanding, li2018measuring}. Although they are often overparameterized, due to the inductive biases of the training algorithms, the intrinsic dimensions of deep neural networks are usually controlled very well, which avoids the curse of dimensionality; (3) Our adopted generative models allow us to efficiently sample from the optimal transport plan. This is very convenient for certain downstream applications such as domain adaptation, where we can generate infinitely many data points paired across domains \citep{liu2016coupled}.

Moreover, the proposed framework can also recover the density of entropy regularized optimal transport plan.
Specifically, we adopt the neural Ordinary Differential Equation (ODE) approach in \citet{chen2018neural} to model the dynamics that how $Z$ gradually evolves to $G(Z)$.  
 We then derive the ODE that describes how the density evolves, and solve the density of the transport plan from the ODE.  The recovery of density requires no extra parameters, and can be evaluated efficiently.

\noindent {\bf Notations}: 
Given a matrix $A\in\RR^{d\times d}$, $\det(A)$ denotes its determinant, $\tr(A)=\sum_{i}A_{ii}$ denotes its trace, $\norm{A}_{\rm F}=\sqrt{\sum_{i, j}A_{ij}^2}$ denotes its Frobenius norm, and $|A|$ denotes a matrix with $[|A|]_{ij}=|A_{ij}|$. We use  $\textrm{dim}(v)$ to denote the dimension of a vector $v$.


\section{Background}
\label{sec:background}

We briefly review some background knowledge on optimal transport and implicit generative learning.

{\bf Optimal Transport}: The idea of optimal transport (OT) originally comes from \citet{monge1781memoire}, which proposes to solve the following problem,
\begin{align}\label{eq:monge}
T^* = \argmin_{T(X)\sim \nu} \mathbb{E}_{X\sim \mu} [c(X, T(X))] ,
\end{align}
where $T(\cdot)$ is a mapping from the space of $\mu$ to the space of $\nu$. The mapping $T^*$ is referred to as \textit{Monge map}, and \eqref{eq:monge} is referred to as Monge formulation of optimal transport. 

Monge formulation, however, is not necessarily feasible. For example, when $X$ is a constant random variable and $Y$ is not, there does not exist such a map $T$ satisfying $T(X)\sim \nu$. The Kantorovich formulation of our interest in  \eqref{eq:intro} is essentially a relaxation of \eqref{eq:monge} by replacing the deterministic mapping with the coupling between $\mu$ and $\nu$. Consequently, \textit{Kantorovich formulation} is guaranteed to be feasible and becomes the classical formulation of optimal transport in existing literature \citep{benamou2015iterative, chizat2015unbalanced, frogner2015learning,solomon2015convolutional, xie2018fast}.

\textbf{Implicit Generative Learning}: For generative learning problems, direct estimation of a probability density function is not always convenient. For example, we may not have enough prior knowledge to specify an appropriate parametric form of the probability density function (pdf). Even when an appropriate parametric pdf is available, computing the maximum likelihood estimator (MLE) can be sometimes neither efficient nor scalable. To address these issues, we resort to implicit generative learning, which do not directly specify the density. Specifically, we consider that the observed variable $X$ is generated by transforming a latent random variable $Z$ (with some known distribution $\rho$) through some unknown mapping $G(\cdot)$, i.e., $X=G(Z)$. We then can train a generative model by estimating $G(\cdot)$ with a properly chosen loss function, which can be easier to compute than MLE. Existing literature usually refer to the distribution of $G(Z)$ as the \textbf{push-forward} of reference distribution $\rho$. Such an implicit generative learning approach also enjoys an additional benefit: We only need to choose $\rho$ that is convenient to sample, e.g., uniform or Gaussian distribution, and we then can generate new samples from our learned distribution directly through the estimated mapping $G$ very efficiently.


For many applications, the target distribution can be quite complicated, in contrast to the distribution $\rho$ being simple. This actually requires the mapping $G$ to be flexible. Therefore, we choose to represent G using deep neural networks (DNNs), which are well known for its universal approximation property, i.e., DNNs with sufficiently many neurons and properly chosen activation functions can approximate any continuous functions over compact support up to an arbitrary error. Early empirical evidence, including variational auto-encoder (VAE, \citet{kingma2013auto}) and generative adversarial networks (GAN, \citet{goodfellow2014generative}) have shown great success of parameterizing $G$ with DNNs. They further motivate a series of variants, which adopt various DNN architectures to learn more complicated generative models \citep{radford2015unsupervised, chen2016infogan,zhao2016energy, dai2017calibrating, jiang2018computation}.



Although the above methods cannot directly estimate the density of the target distribution, for certain applications, we can actually recover the density of $G(Z)$. For example, generative flow methods such as NICE \citep{dinh2014nice}, Real NVP \citep{dinh2016density}, and Glow \citep{kingma2018glow}) impose sparsity constraints on weight matrices, and exploit the hierarchical nature of DNNs to compute the densities layer by layer. Specifically, NICE proposed in \citet{dinh2014nice} denotes the transitions of densities within a neural network as
\begin{align*}
Z \xrightarrow{f_0} h_1 \xrightarrow{f_1}  h_2 \cdots h_{m}\xrightarrow{f_m} G(Z),
\end{align*}
where $h_i$ represents the hidden units of the $i$-th layer and $f_i$ is the transition function. NICE suggest to restrict the Jacobian matrices of $f_i$'s to be triangular. Therefore, $f_i$'s are reversible and the transition of density in each layer can be easily computed. More recently, \citet{chen2018neural} propose a neural ordinary differential equation (neural ODE) approach to compute the transition from $Z$ to $G(Z)$. Specifically, they introduce a dynamical formulation and parameterizing the mapping $G$ using DNNs with recursive structures: They use an ODE to describe how the input $Z$ gradually evolves towards the output $G(Z)$ in continuous time,
\begin{align*}
dz/dt = \xi(z(t),t),
\end{align*}
where $z(t)$ denotes the continuous time interpolation of $Z$, and $\xi(\cdot, \cdot)$ denotes a feedforward-type DNN. Without loss of generality, we choose $z(0) = Z$ and $z(1) = G(Z)$. Then under certain regularity conditions, the mapping $G(\cdot)$ is guaranteed to be reversible, and the density of $G(Z)$ can be computed in $\mathcal{O}(d)$ time, where $d$ is the dimension of $Z$ \citep{grathwohl2018ffjord}.

\begin{wrapfigure}{r}{0.52\textwidth}
  \begin{center}
  \vspace{-20pt}
    \includegraphics[width=0.48\textwidth]{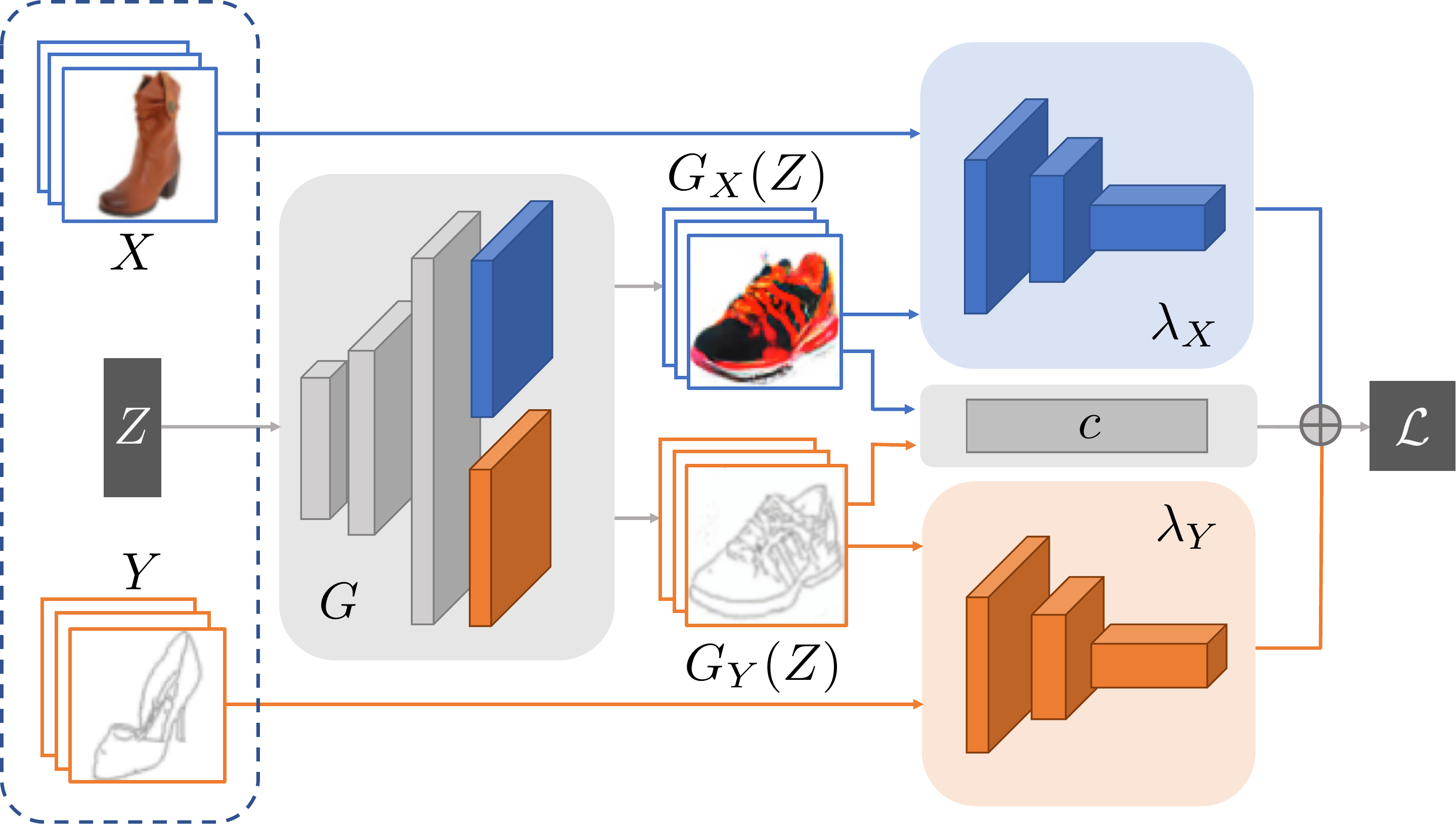}
  \end{center}
  \caption{\label{fig:arche}  An illustration of SPOT. }
  \vspace{-20pt}
\end{wrapfigure}

\section{Scalable OT with Pushforward}\label{sec:framework}

To achieve better efficiency and scalability, we propose a new framework --- named SPOT (Scalable Pushforward of Optimal Transport) --- for solving the optimal transport problem. 
Recall that we aim to find an optimal joint distribution $\gamma$ given by \eqref{eq:intro}. 
Let $\cW_1(X,\mu)$ denotes the standard Wasserstein metric between a random vector $X$ and a distribution $\mu$. Specifically, we write
\begin{align*}
\cW_1(X,\mu) &= \sup_{\lambda_X\in\cF^{1}}\EE_X[\lambda_X(X)] -\mathbb{E}_{U\sim \mu}[\lambda_X(U)],
\end{align*}
where $\cF^{1}$ denotes the class of all $1$-Lipschitz functions from $\RR^d$ to $\RR$. Note that $\cW_1(X,\mu)=0$ indicates $X\sim \mu$. Let $\cW_1(Y,\nu)$ be defined analogously as $\cW_1(X,\mu)$. Then we can rewrite \eqref{eq:intro} as
\begin{align}\label{eq:intro:wasserstein}
\gamma^*= \argmin_{\gamma}~\mathbb{E}_{(X,Y)\sim \gamma} [c(X,Y)], \quad \textrm{subject~to}~\quad \cW_1(X,\mu)=0,~\cW_1(Y,\nu) = 0.
\end{align}
As mentioned earlier, solving $\gamma$ in the space of all continuous distributions is generally intractable. Thus, we adopt the push-forward method, which introduces a mapping $G$ from some latent variable $Z$ to $(X, Y)$. Recall that we denote $(X, Y) = G(Z) = (G_X(Z), G_Y(Z))$ as shown in \eqref{nn-generator}. The latent variable $Z$ follows some distribution $\rho$ that is easy to sample. By the Lagrangian multiplier method and the Kantorovich-Rubinstein duality \citep{villani2008optimal}, we then rewrite \eqref{eq:intro:wasserstein} as
\begin{align}
\min_{G} & \max_{\eta_X,\eta_Y,\lambda_{X}\in\cF^1, \lambda_{Y}\in\cF^{1}}   \mathbb{E}_{Z\sim \rho} [c(G_X (Z),G_Y (Z))] \nonumber \\
& \qquad \quad + \eta_X\mathbb{E}_{Z\sim \rho} [\lambda_X(G_X (Z))] -\mathbb{E}_{U\sim \mu}[\lambda_X(U)]   
 + \eta_Y\mathbb{E}_{Z\sim \rho} [\lambda_Y(G_Y (Z))] -\mathbb{E}_{V\sim \nu}[\lambda_Y(V)]. \label{eq:objective}
\end{align}
Motivated by \citet{arjovsky2017wasserstein}, we then further parameterize $G$, $\lambda_X$, and $\lambda_Y$ with neural networks\footnote{Using a single neural network to parameterize $G$ encourages parameter sharing between $G_X$ and $G_Y$. In fact, we can also parameterize $G_X$ and $G_Y$ with different neural networks.}. We denote $\cG$ as the class of neural networks for parameterizing $G$ and similarly $\cF_X^{1}$ and $\cF_Y^{1}$ as the classes of $1$-Lipschitz functions for $\lambda_X$ and $\lambda_Y$, respectively. 

Since $\cG$, $\cF_X$, and $\cF_Y$ are only finite classes, our parameterization of $G$ cannot exactly represent any continuous distributions of $(X,Y)$ (only up to a small approximation error with sufficiently many neurons). Then the marginal distribution constraints, $G_X(Z) \sim \mu$ and $G_Y(Z) \sim \nu$, are not necessarily satisfied. Therefore, the Lagrangian multipliers can be unbounded and the equilibrium of \eqref{eq:objective} does not necessarily exist. To address this issue, we directly treat $\eta_X=\eta_Y=\eta$ as tuning parameters, and solve the following problem instead
\begin{align}
\min_{G\in \mathcal{G}}  \max_{\lambda_{X}\in \mathcal{F}_{X}^{1}, \lambda_{Y} \in \mathcal{F}_{Y}^{1}} & \mathbb{E}_{Z\sim \rho} [c(G_{X} (Z),G_{Y} (Z))] \notag \\
& + \eta\big(\mathbb{E}_{Z\sim \rho} [\lambda_{X}(G_{X} (Z))] -\mathbb{E}_{X\sim \mu}[\lambda_{X}(X)] 
+ \mathbb{E}_{Z\sim \rho} [\lambda_{Y}(G_{Y} (Z))] -\mathbb{E}_{Y\sim \nu}[\lambda_{Y}(Y)]\big). \label{eq:finalvariant}
\end{align}
We apply alternating stochastic gradient algorithm to solve \eqref{eq:finalvariant}: in each iteration, we perform a few steps of gradient ascent on $\lambda_X$ and $\lambda_Y$, respectively for a fixed $G$, followed by one-step gradient descent on $G$ for fixed $\lambda_X$ and $\lambda_Y$. We use Spectral Normalization (SN, \citet{miyato2018spectral}) to control the Lipschitz constant of $\lambda_X$ and $\lambda_Y$ being smaller than 1. Specifically, SN constrains the spectral norm of each weight matrix $W$ by $\textrm{SN}(W) = W / \sigma(W)$ in every iteration, where $\sigma(W)$ denotes the spectral norm of $W$. Note that $\sigma(W)$ can be efficiently approximated by a simple one-step power method \citep{golub2001eigenvalue}. Therefore, the computationally intensive SVD can be avoided. We summarize the algorithm in Algorithm \ref{alg1} with SN omitted.

\setlength{\textfloatsep}{2pt}
\vspace{-0.1in}
\begin{algorithm}[h]
\caption{Mini-batch Primal Dual Stochastic Gradient Algorithm for SPOT} 
\label{alg1}
\begin{algorithmic} 
    \REQUIRE Datasets $\{x_i\}_{i=1}^N\sim \mu, \{y_j\}_{j=1}^M\sim \nu$; Initialized networks $G$, $\lambda_X$, and $\lambda_Y$ with parameters $w$,  $\theta$, and $\beta$, respectively; $\alpha$, the learning rate; $n_{\rm{critic}}$, the number of gradient ascent for $\lambda_X$ and $\lambda_Y$; $n$, the batch size
    \WHILE{$w$ not converged}
        \FOR{$t=1,2,\cdots,n_{\rm{critic}}$}
        \STATE Sample mini-batch $\{x_i\}_{i=1}^n$ from $\{x_i\}_{i=1}^N$
        \STATE Sample mini-batch $\{y_j\}_{j=1}^n$ from $\{y_j\}_{j=1}^M$
        \STATE Sample mini-batch $\{z_k\}_{k=1}^n$ from $\rho$
        \STATE $g_{\theta} \leftarrow  \nabla_{\theta} (\eta\frac{1}{n}\sum_{k=1}^n \lambda_{X,\theta}(G_{X,w} (z_k))$\\\hspace{3.5cm}$ -\eta\frac{1}{n}\sum_{i=1}^n\lambda_{X,\theta}(x_i) )$
        \STATE $g_{\beta} \leftarrow  \nabla_{\beta} (\eta\frac{1}{n}\sum_{k=1}^n \lambda_{Y,\beta}(G_{Y,w} (z_k))$\\\hspace{3.5cm}$ -\eta\frac{1}{n}\sum_{i=1}^n\lambda_{Y,\beta}(y_i) )$
        \STATE $\theta \leftarrow \theta + \alpha g_{\theta}$, $\beta \leftarrow \beta + \alpha g_{\beta}$
        \ENDFOR
        \STATE Sample mini-batch $\{z_k\}_{k=1}^n$ from $\rho$
        \STATE $g_{w} \leftarrow  \nabla_{w} (\frac{1}{n}\sum_{k=1}^n c(G_{X,w} (z_k),G_{Y,w} (z_k))$
        \\\hspace{2cm}$ + \eta\frac{1}{n}\sum_{k=1}^n \lambda_{X,\theta}(G_{X,w} (z_k)) $
        \\\hspace{3.2cm}$+ \eta\frac{1}{n}\sum_{k=1}^n \lambda_{Y,\beta}(G_{Y,w} (z_k))$
        \STATE $w \leftarrow w + \alpha g_w$
    \ENDWHILE
\end{algorithmic}
\end{algorithm}

{\bf Connection to Wasserstein Generative Adversarial Networks (WGANs)}:
Our proposed framework \eqref{eq:finalvariant} can be viewed as a multi-task learning version of Wasserstein GANs \citep{liu2016coupled, liu2018multi}. Specifically, the mapping $G$ can be viewed as a \textit{generator} that generates samples in the domains $\cX$ and $\cY$. The Lagrangian multipliers $\lambda_X$ and $\lambda_Y$ can be viewed as \textit{discriminators} that evaluate the discrepancies of the generated sample distributions and the target marginal distributions. By restricting $\lambda_X \in \cF_X^1$, $\mathbb{E}_{Z\sim \rho} [\lambda_{X}(G_{X} (Z))] -\mathbb{E}_{X\sim \mu}[\lambda_{X}(X)]$ essentially approximates the Wasserstein distance between the distributions of $G_X(Z)$ and $X$ under the Euclidean ground cost (\citet{villani2008optimal}, the same holds for $Y$). Denote 
\begin{align*}
& \cR(G_X, G_Y) = \mathbb{E}_{Z\sim \rho} [c(G_{X} (Z),G_{Y} (Z))], \quad \textrm{and} 
& d_{w}(G_X, X) = \max_{\lambda_X \in \cF_X^1} \mathbb{E}_{Z\sim \rho} [\lambda_X(G_{X} (Z))] -\mathbb{E}_{X\sim \mu}[\lambda_{X}(X)].
\end{align*}
Let $d_w(G_Y, Y)$ be defined analogously as $d_w(G_X, X)$. We can rewrite \eqref{eq:finalvariant} as
\begin{align}
\label{eq:multi-gan}
\min_{G \in \cG} \eta\big(d_{w}(G_X, X) + d_{w}(G_Y, Y)\big) + \cR(G_X, G_Y),
\end{align}
which essentially learns two Wasserstein GANs with a joint generator $G$ through the regularizer $\cR$. An illustrative example is provided in Figure \ref{fig:arche}.



{\bf Extension to Multiple Marginal Distributions}:
Our proposed framework can be straightforwardly extended to more than two marginal distributions. Consider the ground cost function $c$ taking $m$ inputs $X_1, \dots, X_m$ with $X_i \sim \mu_i$ for $i = 1, \dots, m$. Then the optimal transport problem \eqref{eq:intro} becomes the multi-marginal problem \citep{pass2015multi}:
\begin{align}
\gamma^*= \argmin_{\gamma \in \Pi(\mu_1, \mu_2,\cdots, \mu_m)} \mathbb{E}_{\gamma} [c(X_1,X_2,\cdots, X_m)], \label{eq:multidist}
\end{align}
where $\Pi(\mu_1, \mu_2,\cdots, \mu_m)$ denotes all the joint distributions with marginal distributions satisfying $X_i \sim \mu_i$ for all $i = 1, \dots, m$. Following the same procedure for two distributions, we cast \eqref{eq:multidist} into the following form
\begin{align*}
\min_{G\in \mathcal{G}} & \max_{\lambda_{X_i}\in \mathcal{F}_{X_i}^{\eta}} \mathbb{E}_{Z\sim \rho} [c(G_{X_1} (Z),\cdots, G_{X_m} (Z))]   + \textstyle\sum_{i=1}^m \left(\mathbb{E}_{Z\sim \rho} [\lambda_{X_i}(G_{X_i} (Z))] -\mathbb{E}_{X_i\sim \mu_i}[\lambda_{X_i}(X_i)] \right),
\end{align*}
where $G$ and $\lambda_{X_i}$'s are all parameterized by neural networks. Existing methods for solving the multi-marginal problem \eqref{eq:multidist} suggest to discretize the support of the joint distribution using a refined grid. For complex distributions, the grid size needs to be very large and can be exponential in $m$ \citep{villani2008optimal}. Our parameterization method actually only requires at most $2m$ neural networks, which further corroborates the scalability and efficiency of our framework.



\section{SPOT for Regularized Density Recovery}
\label{section4}

Existing literature has shown that entropy-regularized optimal transportation outperforms the un-regularized counterpart in some applications \citep{erlander1990gravity, cuturi2013sinkhorn}. This is because the entropy regularizer can tradeoff the estimation bias and variance by controlling the smoothness of the density function.

We demonstrate how to efficiently recover the density $p_\gamma$ of the transport plan with entropy regularization.
Instead of parameterizing $G$ by a feedforward neural network, we choose the neural ODE approach, which uses neural networks to approximate the transition from input $Z$ towards output $G(Z)$ in the continuous time. Specifically, we take $z(0) = Z$ and $z(1) = G(Z)$. Let $z(t)$ be the continuous interpolation of $Z$ with density $p(t)$ varying according to time $t$. We split $z(t)$ into $z_1(t)$ and $z_2(t)$ such that $\textrm{dim}(z_1) = \textrm{dim}(X)$ and $\textrm{dim}(z_2) = \textrm{dim}(Y)$. We then write the neural ODE as
\begin{align}
\label{eq:ODE}
dz_1/dt=\xi_1(z(t),t), \quad dz_2/dt=\xi_2(z(t),t),
\end{align}
where $\xi_1$ and $\xi_2$ capture the dynamics of $z(t)$. We parameterize $\xi = (\xi_1, \xi_2)$ by a neural network with parameter $w$. 
We can describe the dynamics of the joint density $p(t)$ in the following proposition.
\begin{proposition}\label{thm:odep}
Let $z$, $z_1$, $z_2$, $\xi_1$ and $\xi_2$ be defined as above. Suppose $\xi_1$ and $\xi_2$ are uniformly Lipschitz continuous in $z$ (the Lipschitz constant is independent of $t$) and continuous in $t$. The log joint density satisfies the following ODE:
\begin{equation}\label{eq:prob}
    \frac{\partial \log p(t)}{\partial t} = - \left(\tr \left(\frac{\partial \xi_1}{\partial z_1}\right)+\tr \left(\frac{\partial \xi_2}{\partial z_2}\right)\right),
\end{equation}
where $\frac{\partial \xi_1}{\partial z_1}$ and $\frac{\partial \xi_2}{\partial z_2}$ are Jacobian matrices of $\xi_1$ and $\xi_2$ with respect to $z_1$ and $z_2$, respectively.
\end{proposition}
Proposition \ref{thm:odep} is a direct result of Theorem 1 in \citet{chen2018neural}. We can now recover the joint density by taking $p_\gamma = p(1)$, which further enables us to efficiently compute the entropy regularizer defined as
\begin{align*}
\cH(p_\gamma) = \EE_{G(Z) \sim \gamma} [\log p_{\gamma}(G(Z))].
\end{align*}
Then we consider the entropy regularized Wasserstein distance $\cL_c(G, \lambda_X, \lambda_Y) + \epsilon \cH(p_\gamma)$ where $\cL_c(G, \lambda_X, \lambda_Y)$ is the objective function in \eqref{eq:finalvariant}. Note that here $G$ is a functional operator of $\xi$, and hence parameterized with $w$. The training algorithm follows Algorithm \ref{alg1}, except that updating $G$ becomes more complex due to involving the neural ODE and the entropy regularizer.

To update $G$, we are essentially updating $w$ using the gradient $g_w = \partial (\cL_c + \epsilon \cH) / \partial w$, where $\epsilon$ is the regularization coefficient. First we compute $\partial \cL_c / \partial w$. We adopt the integral form from \citet{chen2018neural} in the following
\begin{align}
\label{eq:Lw}
\frac{\partial \cL_c}{\partial w} = - \int_0^1 a(t)^\top \frac{\partial \xi(z(t),t)}{\partial w} dt,
\end{align}
where $a(t) = \partial \cL_c / \partial z(t)$ is the so-called ``adjoint variable''. The detailed derivation is slightly involved due to the complicated terms in the chain rule.
We refer the readers to \citet{chen2018neural} for a complete argument. The advantage of introducing $a(t)$ is that we can compute $a(t)$ using the following ODE,
\begin{align*}
\frac{d a(t)}{d t} = -a(t)^\top \frac{\partial \xi(z(t),t)}{\partial z}.
\end{align*}
Then we can use a well developed numerical method to compute \eqref{eq:Lw} efficiently \citep{davis2007methods}.
Next, we compute $\partial \cH / \partial w$ in a similar procedure with $a(t)$ replaced by $b(t) = \partial \cH / \partial \log p(t)$. We then write
\begin{align*}
\frac{\partial \cH}{\partial w} = - \int_0^1 b(t)^\top \frac{\partial \log p(t)}{\partial w} dt.
\end{align*}
Using the same numerical method, we can compute $\partial \cH / \partial w$, which eventually allows us to compute $g_w$ and update $w$.

\vspace{-0.15in}
\section{SPOT for Domain Adaptation}
\vspace{-0.05in}

 Optimal transport has been used in domain adaptation, but existing methods are either computationally inefficient \citep{courty2017joint, damodaran2018deepjdot}, or cannot achieve a state-of-the-art performance \citep{seguy2018large}. Here, we demonstrate that SPOT can tackle large scale domain adaptation problems with state-of-the-art performance.
 
Specifically, we obtain labeled source data $\{x_i\} \sim \mu$, where each data point is associated with a label $v_i$, and target data $\{y_j\} \sim \nu$ with unknown labels. For simplicity, we use $X$ and $Y$ to denote the random vectors following distributions $\mu$ and $\nu$, respectively. The two distributions $\mu$ and $\nu$ can be coupled in a way that each paired samples of $(X,Y)$ from the coupled joint distribution are likely to have the same label. In order to identify such coupling information between source and target data, we propose a new OT-based domain adaptation method --- DASPOT (Domain Adaptation with SPOT) as follows.


Specifically, we jointly train an optimal transport plan and two classifiers for $X$ and $Y$ (denoted by $D_X$ and $D_Y$, respectively). Each classifier is a composition of two neural networks --- an embedding network and a decision network. For simplicity, we denote $D_X = D_{e, X} \circ D_{c, X}$, where $D_{e, X}$ denotes the embedding network, and $D_{c, X}$ denotes the decision network (respectively for $D_Y = D_{e, Y} \circ D_{c, Y}$). We expect the embedding networks to extract high level features of the source and target data, and then find an optimal transport plan to align $X$ and $Y$ based on these high level features using SPOT. Here we choose a ground cost \begin{align}\label{DA-cost}
c(x,y) = \norm{D_{e,X}(x)-D_{e,Y}(y)}^2.
\end{align}
Let $G$ denote the generator of SPOT. The Wasserstein distance of such an OT problem can be written as $\EE_Z\norm{D_{e,X}(G_X(Z))-D_{e,Y}(G_Y(Z))}^2$.


Meanwhile, we train $D_X$ by minimizing the empirical risk $\frac{1}{n}\sum_{i=1}^n[\cE(D_X(x_i),v_i)]$, where $\cE$ denotes the cross entropy loss function, and train $D_Y$ by minimizing 
\begin{align}\label{DA-risk}
\EE_Z[\cE(D_Y(G_Y(Z)),\argmax_{k}[D_X(G_X(Z))]_k)],
\end{align}
where $[v]_k$ denotes the $k$-th entry of the vector $v$. The risk function defined in \eqref{DA-risk} essentially encourages $D_X$ and $D_Y$ to predict each paired (synthetic) samples of $(G_X(Z),G_Y(Z))$ to have the same label.

Eventually, the joint training optimize 
\begin{align*}
\min_{D_X,D_Y,G}\max_{\lambda_X,\lambda_Y}\cL_c(G,\lambda_X,\lambda_Y) + \frac{\eta_{\textrm{s}} }{n}\sum_{i=1}^n[\cE(D_X(x_i),v_i)]
+ \eta_{\rm da}  \EE_Z[\cE(D_Y(G_Y(Z)),\argmax_{k}[D_X(G_X(Z))]_k)],
\end{align*} where $\cL_c(G,\lambda_X,\lambda_Y)$ is the objective function in \eqref{eq:finalvariant} with $c$ defined in \eqref{DA-cost}, and $\eta_{\rm s}, \eta_{\rm da}$ are the tuning parameters. 


\section{Experiments}

We evaluate the SPOT framework on various tasks: Wasserstein distance approximation, density recovery, paired sample generation and domain adaptation. All experiments are implemented with PyTorch using one GTX1080Ti GPU and a Linux desktop computer with 32GB memory, and we adopt the Adam optimizer with configuration parameters 0.5 and 0.999 \citep{kingma2014adam}.

\subsection{Wasserstein Distance (WD) Approximation}
\label{section51}


We first demonstrate that SPOT can accurately and efficiently approximate the Wasserstein distance. We take the Euclidean ground cost, i.e. $c(x, y) = \norm{x - y}$. Then $\EE_{G(Z) \sim \gamma^*}[c(G_X(Z),G_Y(Z))]$ essentially approximates the Wasserstein distance. We take the marginal distributions $\mu$ and $\nu$ as two Gaussian distributions in $\RR^2$ with the same identity covariance matrix. The means are $(-2.5, 0)^\top$ and $(2.5, 0)^\top$, respectively. We find the Wasserstein distance between $\mu$ and $\nu$ equal to 5 by evaluating its closed-form solution. We generate $n = 10^5$ samples from both distributions $\mu$ and $\nu$, respectively. Note that naively applying discretization-based algorithms by dividing the support according to samples requires at least 40 GB memory, which is beyond the memory capability.

We parameterize $G_X$, $G_Y$, $\lambda_X$, and $\lambda_Y$ with fully connected neural networks without sharing parameters. All the networks use the Leaky-ReLU activation \cite{maas2013rectifier}. $G_X$ and $G_Y$ have $2$ hidden layers. $\lambda_X$ and $\lambda_Y$ have $1$ hidden layer. The latent variable $Z$ follows the standard Gaussian distribution in $\RR^2$. We take the batch size equal to 100. 



\textbf{WD vs. Number of Epochs.} We compare the algorithmic behavior of SPOT and Regularized Optimal Transport (ROT, \citet{seguy2017large}) with different regularization coefficients. For SPOT, we set the number of units in each hidden layer equal to 8 and $\eta = 10^4$. 
For ROT, we adopt the code from the authors\footnote{https://github.com/vivienseguy/Large-Scale-OT} with only different input samples, learning rates, and regularization coefficients.

\begin{figure}[h]
\begin{subfigure}[b]{0.27\textwidth}
      \includegraphics[height=0.7\textwidth]{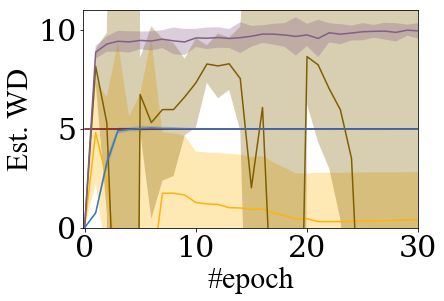}
      \caption{LR =$10^{-3}$}
\end{subfigure}
\begin{subfigure}[b]{0.27\textwidth}
      \includegraphics[height=0.7\textwidth]{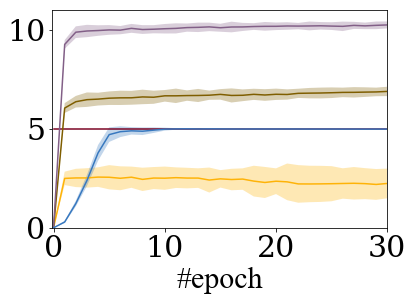}
      \caption{LR =$10^{-4}$}
\end{subfigure}\hspace{-1.8mm}
\begin{subfigure}[b]{0.27\textwidth}
      \includegraphics[height=0.7\textwidth]{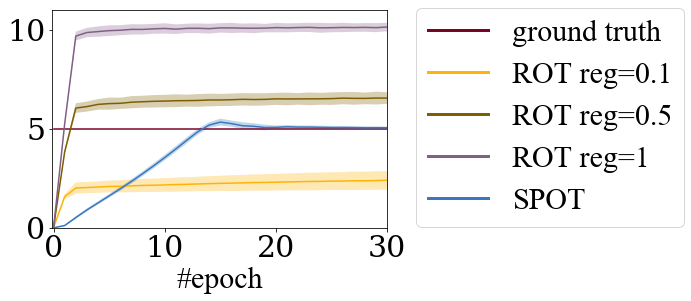}
      \caption{LR =$10^{-5}$}
\end{subfigure}
    \caption{{\label{fig:conv} \it Comparison of convergence between SPOT and ROT. All the curves are averaged over 50 runs with different random seeds, and the shaded areas represent the standard deviation.}}
\end{figure}

Figure \ref{fig:conv} shows the convergence behavior of SPOT and ROT for approximating the Wasserstein distance between $\mu$ and $\nu$ with different learning rates. 
We observe that SPOT converges to the true Wasserstein distance with only 0.6\%, 0.3\%, and 0.3\% relative errors corresponding to Learning Rates (LR) $10^{-3}$, $10^{-4}$, and $10^{-5}$, respectively. In contrast, ROT is very sensitive to its regularization coefficient. Thus, it requires extensive tuning to achieve a good performance.

\begin{wrapfigure}{R}{0.5\textwidth}
  \begin{center}
  \vspace{-20pt}
    \includegraphics[width=0.48\textwidth]{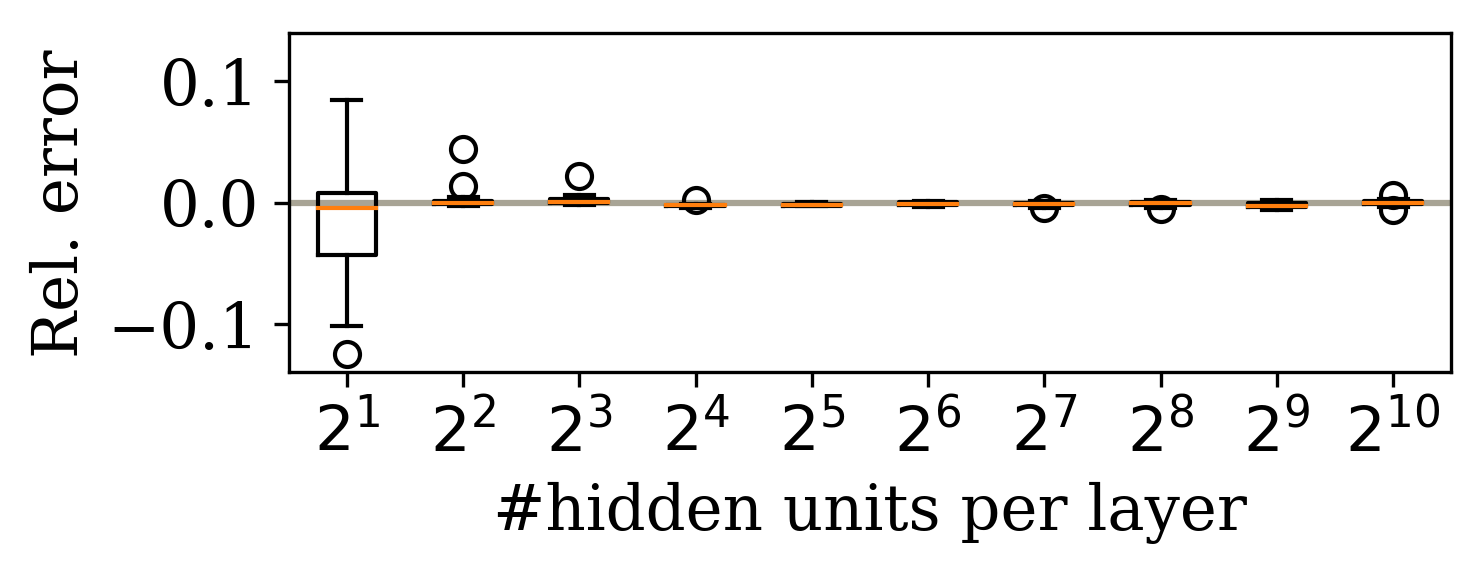}
  \end{center}
  \vspace{-10pt}
  \caption{\label{fig:hid}  Box plots of relative errors of the estimated Wasserstein distance with respect to the number of hidden units per layer. The results are averaged over 50 independent runs. }
  \vspace{-10pt}
\end{wrapfigure}

\textbf{WD vs. Number of Hidden Units.} 
We then explore the adaptivity of SPOT by increasing the network size, while the input data are generated from some low dimensional distribution. Specifically, the number of hidden units per layer varies from $2$ to $2^{10}$. Recall that we parameterize $G$ with two 2-hidden-layer neural networks, and $\lambda_X$, $\lambda_Y$ with two 1-hidden-layer neural networks. Accordingly, the number of parameters in $G$ varies from $36$ to about $2\times10^6$, and that in $\lambda_X$ or $\lambda_Y$ varies from $12$ to about $2,000$.  The tuning parameter $\eta$ also varies corresponding to the number of hidden units in $\lambda_X$, $\lambda_Y$. We use $\eta=10^{5}$ for $2^1, 2^2$ and $2^3$ hidden units per layer, $\eta=2\times 10^{4}$ for $2^4, 2^5$ and $2^6$ hidden units per layer, $\eta=10^{4}$ for $2^7$ and $2^8$ hidden units per layer, $\eta=2\times 10^{3}$ for $2^9$, and $2^{10}$ hidden units per layer.


Figure \ref{fig:hid} shows the estimated WD with respect to the number of hidden units per layer. For large neural networks that have $2^9$ or $2^{10}$ hidden units per layer, i.e., $5.2\times 10^{5}$ or $2.0\times10^{6}$ parameters, the number of parameters is far larger than the number of samples. Therefore, the model is heavily overparameterized. As we can observe in Figure \ref{fig:hid}, the relative error however, does not increase as the number of parameters grows. This suggests that SPOT is quite robust with respect to the network size.


\subsection{Density Recovery}

We demonstrate that SPOT can effectively recover the joint density with entropy regularization. We adopt the neural ODE approach as described in Section \ref{section4}. Denote $\phi(a,b)$ as the density of the Gaussian distribution $N(a, b)$. We take the marginal distributions $\mu$ and $\nu$ as (1) Gaussian distributions $\phi(0,1)$ and $\phi(2,0.5)$; (2) mixtures of Gaussian $\frac{1}{2}\phi(-1,0.5)+\frac{1}{2}\phi(1,0.5)$ and $\frac{1}{2}\phi(-2,0.5)+\frac{1}{2}\phi(2,0.5)$.
The ground cost is the Euclidean square function, i.e., $c(x, y) = \norm{x - y}^2$. We run the training algorithm for $6 \times 10^5$ iterations and in each iteration, we generate 500 samples from $\mu$ and $\nu$, respectively. We parameterize $\xi$ with a 3-hidden-layer fully-connected neural network with 64 hidden units per layer, and the latent dimension is 2.  We take $\eta=10^6$.

\begin{wrapfigure}{R}{0.55\textwidth}
  \begin{center}
  \vspace{-10pt}
    \includegraphics[width=0.53\textwidth]{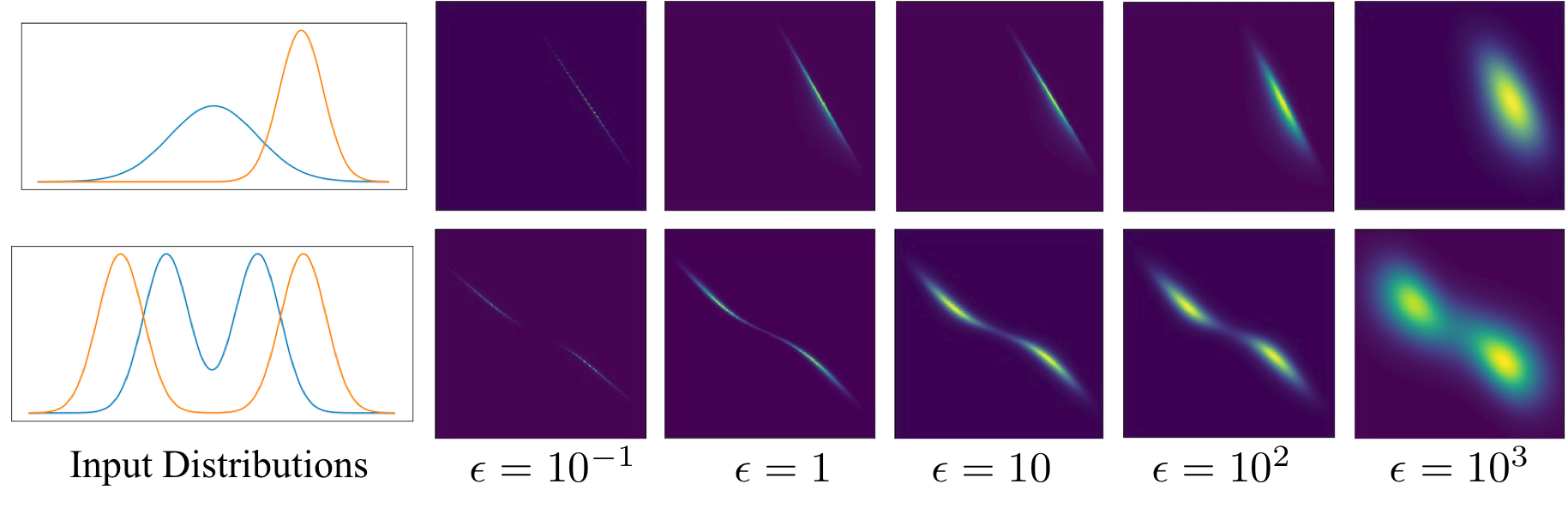}
  \end{center}
  \caption{\label{fig:node}  Visualization of the marginal distributions and the joint density of the optimal transport plan. }
\end{wrapfigure}

Figure \ref{fig:node} shows the input marginal densities and heat maps of output joint densities. We can see that a larger regularization coefficient $\epsilon$ yields a smoother joint density for the optimal transport plan. Note that with continuous marginal distributions and the Euclidean square ground cost, the joint density of the unregularized optimal transport degenerates to a generalized impulse function (i.e., a generalized Dirac $\delta$ function that has nonzero value on a manifold instead of one atom, as shown in \citet{rachev1985monge, onural2006impulse}). Entropy regularization prevents such degeneracy by enforcing smoothness of the density.



\subsection{Sample Generation}
\label{sec:section53}

\begin{figure*}[t!]
\centering
\includegraphics[width=0.99\textwidth]{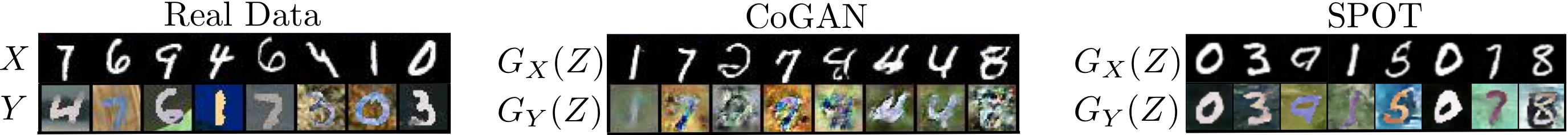}
\caption{\label{fig:generated_mnist} \it Generated samples of SPOT and CoGAN on the MNIST-MNISTM task.}
\end{figure*}

We show that SPOT can generate paired samples $(G_X(Z), G_Y(Z))$ from unpaired data $X$ and $Y$ that are sampled from marginal distributions $\mu$ and $\nu$, respectively.


\begin{wrapfigure}{r}{0.45\textwidth}
  \begin{center}
    \includegraphics[width=0.43\textwidth]{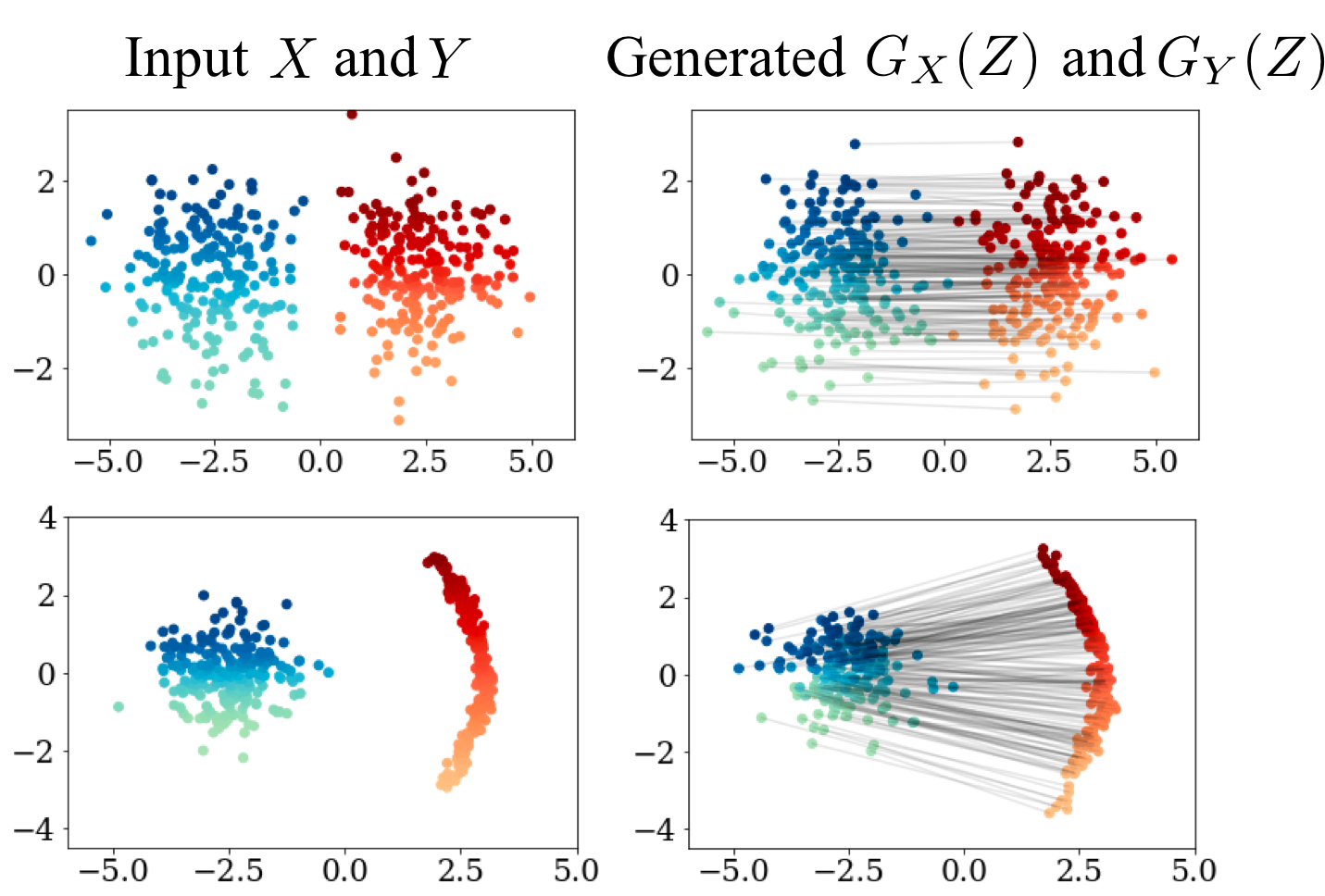}
  \end{center}
  \caption{\label{fig:synthetic}  Visualization of input samples and generated samples. The black lines represent the paired relation. }
\end{wrapfigure}

\textbf{Synthetic Data.} 
We take the squared Euclidean cost, i.e. $c(x, y) = \norm{x - y}^2$, and adopt the same implementation and sample size as in Section \ref{section51} with learning rate $10^{-3}$ and 32 hidden units per layer. 
Figure \ref{fig:synthetic} illustrates the input samples and the generated samples with two sets of different marginal distributions: The upper row corresponds to the same Gaussian distributions as in Section \ref{section51}. The lower row takes $X$ as Gaussian distribution with mean $(-2.5,0)^\top$ and covariance $0.5I$, $Y$ as $(\sin (Y_1)+{Y}_2, 2Y_1-3)^\top$, where $Y_1$ follows a uniform distribution on $[0,3]$, and $Y_2$ follows a Gaussian distribution $N(2, 0.1)$. 

We observe that the generated samples and the input samples are approximately identically distributed. Additionally, the paired relationship is as expected -- the upper mass is transported to the upper region, and the lower mass is transported to the lower region. 

\textbf{Real Data.} 
We next show SPOT is able to generate high quality paired samples from two unpaired real datasets: MNIST \citep{lecun1998gradient} and MNISTM \citep{ganin2014unsupervised}. The handwritten digits in MNIST and MNISTM datasets have different backgrounds and foregrounds (see Figrue \ref{fig:generated_mnist}). The digits in paired images however, are expected to have similar contours. We leverage this prior knowledge\footnote{For OT problems, $c$ can be viewed as a way to add prior knowledge to the problem \citep{peyre2017computational}.} by adopting a semantic-aware cost function \citep{li2018semantic} to extract the edge of handwritten letters, i.e., we use the following cost function
\begin{align*}
c(x,y)= \textstyle\sum_{i = 1}^2 \sum_{j = 1}^3 \left\lVert |C_i * x_j| - |C_i * y_j|\right\rVert_\textrm{F},
\end{align*}
where $C_1$ and $C_2$ denote the Sobel filter \citep{sobel1990isotropic}, and $x_j$'s and $y_j$'s are the three channels of RGB images. The operator $*$ denotes the matrix convolution. We set
\begin{align*}
C_1 = \begin{bmatrix}
-1&0&1\\
-2&0&2\\
-1&0&1
\end{bmatrix} ~~\textrm{and}~~
C_2 = \begin{bmatrix}
1&2&1\\
0&0&0\\
-1&-2&-1
\end{bmatrix},
\end{align*}
with $C_1$ and $C_2$ defining two extraction directions.

We now use separate neural networks to parameterize $G_X$ and $G_Y$ instead of taking $G_X$ and $G_Y$ as outputs of a common network. Note that $G_X$ and $G_Y$ does not share parameters. Specifically,
we use two 4-layer convolutional layers in each neural network for $G_X$ or $G_Y$, and two 5-layer convolutional neural networks for $\lambda_{X}$ and $\lambda_{Y}$. More detailed network settings are provided in Appendix~\ref{sec:cnn_structure}. The batch size is 32, and we train the framework with $2\times 10^5$ iterations until the generated samples become stable.

Figure \ref{fig:generated_mnist} shows the generated samples of SPOT. We also reproduce the results of CoGAN with the code from the authors\footnote{https://github.com/mingyuliutw/CoGAN}. As can be seen, with approximately the same network size, SPOT yields paired images with better quality than CoGAN: The contours of the paired results of SPOT are nearly identical, while the results of CoGAN have no clear paired relation. Besides, the images corresponding to $G_Y(Z)$ in SPOT have colorful foreground and background, while in CoGAN there are only few colors. Recall that in SPOT, the paired relation is encouraged by ground cost $c$, and in CoGAN it is encouraged by sharing parameters. By leveraging prior knowledge in ground cost $c$, the paired relation is more accurately controlled without compromising the quality of the generated images.



We further test our framework on more complex real datasets: Photo-Monet dataset \cite{zhu2017unpaired} and Edges-Shoes dataset \cite{isola2017image}. We adopt the Euclidean cost function for Photo-Monet dataset, and the semantic-aware cost function as in MNIST-MNISTM for Edges-Shoes dataset. Other implementations remain the same as the MNIST-MINSTM experiment. 

\begin{figure}[t!]
\centering
\includegraphics[width=0.98\textwidth]{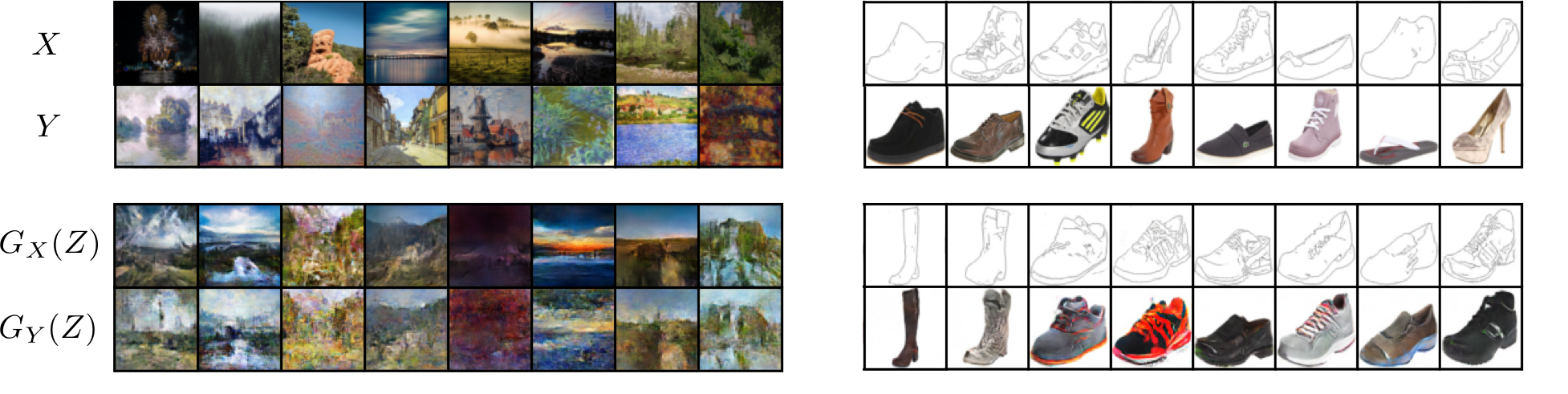}
\caption{\label{fig:generated_real} \it Generated samples of SPOT on Photos-Monet and Sketches-Shoes datasets.}
\end{figure}

Figure \ref{fig:generated_real} demonstrates the generated samples of both datasets. We observe that the generated images have a desired paired relation: For each $Z$, $G_X(Z)$ and $G_Y(Z)$ gives a pair of corresponding scenery and shoe. The generated images are also of high quality, especially considering that Photo-Monet dataset is a pretty small but complex dataset with 6,288 photos and 1,073 paintings. 


\subsection{Domain Adaptation}\label{sec:DA}
We choose $\eta_{\textrm{s}} = 10^3$ for all experiments. We set $\eta_{\textrm{da}}=0$ for the first $10^5$ iteration to wait the generators to be well trained. Then we set $\eta_{\textrm{da}}=10$ for the next $3\times 10^5$ iteration. We take totally $4\times 10^5$ iterations, and set the learning rate equal to $10^{-4}$ and batch size equal to $128$ for all experiments. 

We evaluate DASPOT with the MNIST, MNISTM, USPS \citep{hull1994database}, and SVHN \citep{netzer2011reading} datasets. We denote a domain adaptation task as Source Domain $\rightarrow$ Target Domain. 
For the tasks MNIST $\rightarrow$ USPS, USPS $\rightarrow$ MNIST and MNIST $\rightarrow$ MNISTM, we use three 4-layer networks for $D,\lambda_X$,and $\lambda_Y$, and two 5-layer networks for $G_X$ and $G_Y$. For the task SVHN $\rightarrow$ MNIST, we use three 5-layer downsampling ResNets \cite{he2016deep} for $D,\lambda_X,$ and $\lambda_Y$, and two 5-layer upsampling ResNets for $G_X$ and $G_Y$. More detailed implementations are provided in Appendices~\ref{sec:cnn_structure} and \ref{sec:resnet_structure}.

\begin{wraptable}{r}{0.6\linewidth}
\caption{\it Domain Adaptation Experiments on multiple tasks.}
\label{tab:DAexp}
\scalebox{1.2}
{\footnotesize
\begin{tabular}{l|l|l|l|l}
		\hline
		Source		&  MNIST	&  USPS 	&  SVHN 	&  MNIST\\ 
		Target		&  USPS 	&  MNIST	&  MNIST	&  MNISTM\\ \hline
		ROT 	&  $72.6\%$	&  $60.5\%$	&  $62.9\%$	&  $-$ \\ \hline
		StochJDOT 	&  $93.6\%$	&  $90.5\%$	&  $67.6\%$	&  $66.7\%$ \\ \hline
		DeepJDOT 	&  $95.7\%$	&  $96.4\%$	&  $\textbf{96.7\%}$	&  $92.4\%$ \\ \hline
		DASPOT 		&  $\textbf{97.5\%}$	&  $\textbf{96.5\%}$	&  $96.2\%$	&  $\textbf{94.9\%}$ \\ \hline
\end{tabular}}
\end{wraptable} 

We compare the performance of DASPOT with other optimal transport based domain adaptation methods: ROT \citep{seguy2018large}, StochJDOT \citep{damodaran2018deepjdot} and DeepJDOT \citep{damodaran2018deepjdot}. As can be seen in Table \ref{tab:DAexp}, DASPOT achieves equal or better performances on all the tasks.

%

Moreover, we show that DeepJDOT is not as efficient as DASPOT. 
For example, in the MNIST $\rightarrow$ USPS task, DASPOT requires $169$s running time to achieve a $95\%$ accuracy, while DeepJDOT requires $518$s running time to achieve the same accuracy. The reason behind is that DeepJDOT needs to solve a series of optimal transport problems using Sinkhorn algorithm. The implementation of DeepJDOT is adapted from the authors' code\footnote{https://github.com/bbdamodaran/deepJDOT}.



\section{Discussion}\label{sec:discussion}

Existing literature shows that several stochastic algorithms can efficiently compute the Wasserstein distance between two continuous distributions. These algorithms, however, only apply to the dual of the OT problem \eqref{eq:intro}, and cannot provide the optimal transport plan. For example, \citet{genevay2016stochastic} suggest to expand the dual variables in two reproducing kernel Hilbert spaces. They then apply the Stochastic Averaged Gradient (SAG) algorithm to compute the optimal objective value of OT with continuous marginal distributions or semi-discrete marginal distributions (i.e., one marginal distribution is continuous and the other is discrete). The follow-up work, \citet{seguy2017large}, parameterize the dual variables with neural networks and apply the Stochastic Gradient Descent (SGD) algorithm to eventually achieve a better convergence. These two methods can only provide the optimal transport plan and recover the joint density when the densities of the marginal distributions are known. This is prohibitive in most applications, since we only have access to the empirical data. Our framework actually allows us to efficiently compute the joint density from the transformation of the latent variable $Z$ as in Section \ref{section4}.

\section*{Acknowledgement}

This work is partially supported by the grant NSF IIS 1717916 and NSF CMMI 1745382. We thank Dr. Bharath Bhushan Damodaran for his timely help about the implementation of DeepJDOT method.

\bibliography{example_paper}

\begin{thebibliography}{57}
\expandafter\ifx\csname natexlab\endcsname\relax\def\natexlab#1{#1}\fi
\expandafter\ifx\csname url\endcsname\relax
  \def\url#1{\texttt{#1}}\fi
\expandafter\ifx\csname urlprefix\endcsname\relax\def\urlprefix{}\fi

\bibitem[{Arjovsky et~al.(2017)Arjovsky, Chintala and
  Bottou}]{arjovsky2017wasserstein}
\textsc{Arjovsky, M.}, \textsc{Chintala, S.} and \textsc{Bottou, L.} (2017).
\newblock Wasserstein gan.
\newblock \textit{arXiv preprint arXiv:1701.07875}.

\bibitem[{Benamou et~al.(2015)Benamou, Carlier, Cuturi, Nenna and
  Peyr{\'e}}]{benamou2015iterative}
\textsc{Benamou, J.-D.}, \textsc{Carlier, G.}, \textsc{Cuturi, M.},
  \textsc{Nenna, L.} and \textsc{Peyr{\'e}, G.} (2015).
\newblock Iterative bregman projections for regularized transportation
  problems.
\newblock \textit{SIAM Journal on Scientific Computing}, \textbf{37}
  A1111--A1138.

\bibitem[{Brock et~al.(2018)Brock, Donahue and Simonyan}]{brock2018large}
\textsc{Brock, A.}, \textsc{Donahue, J.} and \textsc{Simonyan, K.} (2018).
\newblock Large scale gan training for high fidelity natural image synthesis.
\newblock \textit{arXiv preprint arXiv:1809.11096}.

\bibitem[{Carlier(2012)}]{carlier2012optimal}
\textsc{Carlier, G.} (2012).
\newblock Optimal transportation and economic applications.
\newblock \textit{Lecture Notes.(Cited on page 2.)}.

\bibitem[{Chen et~al.(2018)Chen, Rubanova, Bettencourt and
  Duvenaud}]{chen2018neural}
\textsc{Chen, T.~Q.}, \textsc{Rubanova, Y.}, \textsc{Bettencourt, J.} and
  \textsc{Duvenaud, D.} (2018).
\newblock Neural ordinary differential equations.
\newblock \textit{arXiv preprint arXiv:1806.07366}.

\bibitem[{Chen et~al.(2016)Chen, Duan, Houthooft, Schulman, Sutskever and
  Abbeel}]{chen2016infogan}
\textsc{Chen, X.}, \textsc{Duan, Y.}, \textsc{Houthooft, R.}, \textsc{Schulman,
  J.}, \textsc{Sutskever, I.} and \textsc{Abbeel, P.} (2016).
\newblock Infogan: Interpretable representation learning by information
  maximizing generative adversarial nets.
\newblock In \textit{Advances in neural information processing systems}.

\bibitem[{Chizat et~al.(2015)Chizat, Peyr{\'e}, Schmitzer and
  Vialard}]{chizat2015unbalanced}
\textsc{Chizat, L.}, \textsc{Peyr{\'e}, G.}, \textsc{Schmitzer, B.} and
  \textsc{Vialard, F.-X.} (2015).
\newblock Unbalanced optimal transport: geometry and kantorovich formulation.
\newblock \textit{arXiv preprint arXiv:1508.05216}.

\bibitem[{Courty et~al.(2017{\natexlab{a}})Courty, Flamary, Habrard and
  Rakotomamonjy}]{courty2017joint}
\textsc{Courty, N.}, \textsc{Flamary, R.}, \textsc{Habrard, A.} and
  \textsc{Rakotomamonjy, A.} (2017{\natexlab{a}}).
\newblock Joint distribution optimal transportation for domain adaptation.
\newblock In \textit{Advances in Neural Information Processing Systems}.

\bibitem[{Courty et~al.(2017{\natexlab{b}})Courty, Flamary, Tuia and
  Rakotomamonjy}]{courty2017optimal}
\textsc{Courty, N.}, \textsc{Flamary, R.}, \textsc{Tuia, D.} and
  \textsc{Rakotomamonjy, A.} (2017{\natexlab{b}}).
\newblock Optimal transport for domain adaptation.
\newblock \textit{IEEE transactions on pattern analysis and machine
  intelligence}, \textbf{39} 1853--1865.

\bibitem[{Cuturi(2013)}]{cuturi2013sinkhorn}
\textsc{Cuturi, M.} (2013).
\newblock Sinkhorn distances: Lightspeed computation of optimal transport.
\newblock In \textit{Advances in neural information processing systems}.

\bibitem[{Dai et~al.(2017)Dai, Almahairi, Bachman, Hovy and
  Courville}]{dai2017calibrating}
\textsc{Dai, Z.}, \textsc{Almahairi, A.}, \textsc{Bachman, P.}, \textsc{Hovy,
  E.} and \textsc{Courville, A.} (2017).
\newblock Calibrating energy-based generative adversarial networks.
\newblock \textit{arXiv preprint arXiv:1702.01691}.

\bibitem[{Damodaran et~al.(2018)Damodaran, Kellenberger, Flamary, Tuia and
  Courty}]{damodaran2018deepjdot}
\textsc{Damodaran, B.~B.}, \textsc{Kellenberger, B.}, \textsc{Flamary, R.},
  \textsc{Tuia, D.} and \textsc{Courty, N.} (2018).
\newblock Deepjdot: Deep joint distribution optimal transport for unsupervised
  domain adaptation.
\newblock \textit{arXiv preprint arXiv:1803.10081}.

\bibitem[{Davis and Rabinowitz(2007)}]{davis2007methods}
\textsc{Davis, P.~J.} and \textsc{Rabinowitz, P.} (2007).
\newblock \textit{Methods of numerical integration}.
\newblock Courier Corporation.

\bibitem[{Dinh et~al.(2014)Dinh, Krueger and Bengio}]{dinh2014nice}
\textsc{Dinh, L.}, \textsc{Krueger, D.} and \textsc{Bengio, Y.} (2014).
\newblock Nice: Non-linear independent components estimation.
\newblock \textit{arXiv preprint arXiv:1410.8516}.

\bibitem[{Dinh et~al.(2016)Dinh, Sohl-Dickstein and Bengio}]{dinh2016density}
\textsc{Dinh, L.}, \textsc{Sohl-Dickstein, J.} and \textsc{Bengio, S.} (2016).
\newblock Density estimation using real nvp.
\newblock \textit{arXiv preprint arXiv:1605.08803}.

\bibitem[{Erlander and Stewart(1990)}]{erlander1990gravity}
\textsc{Erlander, S.} and \textsc{Stewart, N.~F.} (1990).
\newblock \textit{The gravity model in transportation analysis: theory and
  extensions}, vol.~3.
\newblock Vsp.

\bibitem[{Frogner et~al.(2015)Frogner, Zhang, Mobahi, Araya and
  Poggio}]{frogner2015learning}
\textsc{Frogner, C.}, \textsc{Zhang, C.}, \textsc{Mobahi, H.}, \textsc{Araya,
  M.} and \textsc{Poggio, T.~A.} (2015).
\newblock Learning with a wasserstein loss.
\newblock In \textit{Advances in Neural Information Processing Systems}.

\bibitem[{Galichon(2017)}]{galichon2017survey}
\textsc{Galichon, A.} (2017).
\newblock A survey of some recent applications of optimal transport methods to
  econometrics.
\newblock \textit{The Econometrics Journal}, \textbf{20} C1--C11.

\bibitem[{Ganin and Lempitsky(2014)}]{ganin2014unsupervised}
\textsc{Ganin, Y.} and \textsc{Lempitsky, V.} (2014).
\newblock Unsupervised domain adaptation by backpropagation.
\newblock \textit{arXiv preprint arXiv:1409.7495}.

\bibitem[{Genevay et~al.(2016)Genevay, Cuturi, Peyr{\'e} and
  Bach}]{genevay2016stochastic}
\textsc{Genevay, A.}, \textsc{Cuturi, M.}, \textsc{Peyr{\'e}, G.} and
  \textsc{Bach, F.} (2016).
\newblock Stochastic optimization for large-scale optimal transport.
\newblock In \textit{Advances in Neural Information Processing Systems}.

\bibitem[{Golub and Van~der Vorst(2001)}]{golub2001eigenvalue}
\textsc{Golub, G.~H.} and \textsc{Van~der Vorst, H.~A.} (2001).
\newblock Eigenvalue computation in the 20th century.
\newblock In \textit{Numerical analysis: historical developments in the 20th
  century}. Elsevier, 209--239.

\bibitem[{Goodfellow et~al.(2014)Goodfellow, Pouget-Abadie, Mirza, Xu,
  Warde-Farley, Ozair, Courville and Bengio}]{goodfellow2014generative}
\textsc{Goodfellow, I.}, \textsc{Pouget-Abadie, J.}, \textsc{Mirza, M.},
  \textsc{Xu, B.}, \textsc{Warde-Farley, D.}, \textsc{Ozair, S.},
  \textsc{Courville, A.} and \textsc{Bengio, Y.} (2014).
\newblock Generative adversarial nets.
\newblock In \textit{Advances in neural information processing systems}.

\bibitem[{Grathwohl et~al.(2018)Grathwohl, Chen, Betterncourt, Sutskever and
  Duvenaud}]{grathwohl2018ffjord}
\textsc{Grathwohl, W.}, \textsc{Chen, R.~T.}, \textsc{Betterncourt, J.},
  \textsc{Sutskever, I.} and \textsc{Duvenaud, D.} (2018).
\newblock Ffjord: Free-form continuous dynamics for scalable reversible
  generative models.
\newblock \textit{arXiv preprint arXiv:1810.01367}.

\bibitem[{Gross et~al.(2016)Gross, Wan, Rasch, Caldwell, Williamson, Klocke,
  Jablonowski, Thatcher, Wood, Cullen et~al.}]{gross2016recent}
\textsc{Gross, M.}, \textsc{Wan, H.}, \textsc{Rasch, P.~J.}, \textsc{Caldwell,
  P.~M.}, \textsc{Williamson, D.~L.}, \textsc{Klocke, D.}, \textsc{Jablonowski,
  C.}, \textsc{Thatcher, D.~R.}, \textsc{Wood, N.}, \textsc{Cullen, M.}
  \textsc{et~al.} (2016).
\newblock Recent progress and review of issues related to physics dynamics
  coupling in geophysical models.
\newblock \textit{arXiv preprint arXiv:1605.06480}.

\bibitem[{He et~al.(2015)He, Zhang, Ren and Sun}]{he2015delving}
\textsc{He, K.}, \textsc{Zhang, X.}, \textsc{Ren, S.} and \textsc{Sun, J.}
  (2015).
\newblock Delving deep into rectifiers: Surpassing human-level performance on
  imagenet classification.
\newblock In \textit{Proceedings of the IEEE international conference on
  computer vision}.

\bibitem[{He et~al.(2016)He, Zhang, Ren and Sun}]{he2016deep}
\textsc{He, K.}, \textsc{Zhang, X.}, \textsc{Ren, S.} and \textsc{Sun, J.}
  (2016).
\newblock Deep residual learning for image recognition.
\newblock In \textit{Proceedings of the IEEE conference on computer vision and
  pattern recognition}.

\bibitem[{Hull(1994)}]{hull1994database}
\textsc{Hull, J.~J.} (1994).
\newblock A database for handwritten text recognition research.
\newblock \textit{IEEE Transactions on pattern analysis and machine
  intelligence}, \textbf{16} 550--554.

\bibitem[{Isola et~al.(2017)Isola, Zhu, Zhou and Efros}]{isola2017image}
\textsc{Isola, P.}, \textsc{Zhu, J.-Y.}, \textsc{Zhou, T.} and \textsc{Efros,
  A.~A.} (2017).
\newblock Image-to-image translation with conditional adversarial networks.
\newblock \textit{arXiv preprint}.

\bibitem[{Jiang et~al.(2018)Jiang, Chen, Chen, Liu, Wang and
  Zhao}]{jiang2018computation}
\textsc{Jiang, H.}, \textsc{Chen, Z.}, \textsc{Chen, M.}, \textsc{Liu, F.},
  \textsc{Wang, D.} and \textsc{Zhao, T.} (2018).
\newblock On computation and generalization of gans with spectrum control.
\newblock \textit{arXiv preprint arXiv:1812.10912}.

\bibitem[{Kingma and Ba(2014)}]{kingma2014adam}
\textsc{Kingma, D.~P.} and \textsc{Ba, J.} (2014).
\newblock Adam: A method for stochastic optimization.
\newblock \textit{arXiv preprint arXiv:1412.6980}.

\bibitem[{Kingma and Dhariwal(2018)}]{kingma2018glow}
\textsc{Kingma, D.~P.} and \textsc{Dhariwal, P.} (2018).
\newblock Glow: Generative flow with invertible 1x1 convolutions.
\newblock In \textit{Advances in Neural Information Processing Systems}.

\bibitem[{Kingma and Welling(2013)}]{kingma2013auto}
\textsc{Kingma, D.~P.} and \textsc{Welling, M.} (2013).
\newblock Auto-encoding variational bayes.
\newblock \textit{arXiv preprint arXiv:1312.6114}.

\bibitem[{LeCun et~al.(1998)LeCun, Bottou, Bengio and
  Haffner}]{lecun1998gradient}
\textsc{LeCun, Y.}, \textsc{Bottou, L.}, \textsc{Bengio, Y.} and
  \textsc{Haffner, P.} (1998).
\newblock Gradient-based learning applied to document recognition.
\newblock \textit{Proceedings of the IEEE}, \textbf{86} 2278--2324.

\bibitem[{Li et~al.(2018{\natexlab{a}})Li, Farkhoor, Liu and
  Yosinski}]{li2018measuring}
\textsc{Li, C.}, \textsc{Farkhoor, H.}, \textsc{Liu, R.} and \textsc{Yosinski,
  J.} (2018{\natexlab{a}}).
\newblock Measuring the intrinsic dimension of objective landscapes.
\newblock \textit{arXiv preprint arXiv:1804.08838}.

\bibitem[{Li et~al.(2018{\natexlab{b}})Li, Liang, Jia and
  Xing}]{li2018semantic}
\textsc{Li, P.}, \textsc{Liang, X.}, \textsc{Jia, D.} and \textsc{Xing, E.~P.}
  (2018{\natexlab{b}}).
\newblock Semantic-aware grad-gan for virtual-to-real urban scene adaption.
\newblock \textit{arXiv preprint arXiv:1801.01726}.

\bibitem[{Liu and Tuzel(2016)}]{liu2016coupled}
\textsc{Liu, M.-Y.} and \textsc{Tuzel, O.} (2016).
\newblock Coupled generative adversarial networks.
\newblock In \textit{Advances in neural information processing systems}.

\bibitem[{Liu et~al.(2018)Liu, Wang, Jin and Wassell}]{liu2018multi}
\textsc{Liu, Y.}, \textsc{Wang, Z.}, \textsc{Jin, H.} and \textsc{Wassell, I.}
  (2018).
\newblock Multi-task adversarial network for disentangled feature learning.
\newblock In \textit{Proceedings of the IEEE Conference on Computer Vision and
  Pattern Recognition}.

\bibitem[{Maas et~al.(2013)Maas, Hannun and Ng}]{maas2013rectifier}
\textsc{Maas, A.~L.}, \textsc{Hannun, A.~Y.} and \textsc{Ng, A.~Y.} (2013).
\newblock Rectifier nonlinearities improve neural network acoustic models.
\newblock In \textit{Proc. icml}, vol.~30.

\bibitem[{Miyato et~al.(2018)Miyato, Kataoka, Koyama and
  Yoshida}]{miyato2018spectral}
\textsc{Miyato, T.}, \textsc{Kataoka, T.}, \textsc{Koyama, M.} and
  \textsc{Yoshida, Y.} (2018).
\newblock Spectral normalization for generative adversarial networks.
\newblock \textit{arXiv preprint arXiv:1802.05957}.

\bibitem[{Monge(1781)}]{monge1781memoire}
\textsc{Monge, G.} (1781).
\newblock M{\'e}moire sur la th{\'e}orie des d{\'e}blais et des remblais.
\newblock \textit{Histoire de l'Acad{\'e}mie Royale des Sciences de Paris}.

\bibitem[{Netzer et~al.(2011)Netzer, Wang, Coates, Bissacco, Wu and
  Ng}]{netzer2011reading}
\textsc{Netzer, Y.}, \textsc{Wang, T.}, \textsc{Coates, A.}, \textsc{Bissacco,
  A.}, \textsc{Wu, B.} and \textsc{Ng, A.~Y.} (2011).
\newblock Reading digits in natural images with unsupervised feature learning.
\newblock In \textit{NIPS workshop on deep learning and unsupervised feature
  learning}, vol. 2011.

\bibitem[{Onural(2006)}]{onural2006impulse}
\textsc{Onural, L.} (2006).
\newblock Impulse functions over curves and surfaces and their applications to
  diffraction.
\newblock \textit{Journal of mathematical analysis and applications},
  \textbf{322} 18--27.

\bibitem[{Pass(2015)}]{pass2015multi}
\textsc{Pass, B.} (2015).
\newblock Multi-marginal optimal transport: theory and applications.
\newblock \textit{ESAIM: Mathematical Modelling and Numerical Analysis},
  \textbf{49} 1771--1790.

\bibitem[{Peyr{\'e} et~al.(2017)Peyr{\'e}, Cuturi
  et~al.}]{peyre2017computational}
\textsc{Peyr{\'e}, G.}, \textsc{Cuturi, M.} \textsc{et~al.} (2017).
\newblock Computational optimal transport.
\newblock Tech. rep.

\bibitem[{Rachev(1985)}]{rachev1985monge}
\textsc{Rachev, S.~T.} (1985).
\newblock The monge--kantorovich mass transference problem and its stochastic
  applications.
\newblock \textit{Theory of Probability \& Its Applications}, \textbf{29}
  647--676.

\bibitem[{Radford et~al.(2015)Radford, Metz and
  Chintala}]{radford2015unsupervised}
\textsc{Radford, A.}, \textsc{Metz, L.} and \textsc{Chintala, S.} (2015).
\newblock Unsupervised representation learning with deep convolutional
  generative adversarial networks.
\newblock \textit{arXiv preprint arXiv:1511.06434}.

\bibitem[{Santambrogio(2010)}]{santambrogio2010models}
\textsc{Santambrogio, F.} (2010).
\newblock Models and applications of optimal transport in economics, traffic
  and urban planning.
\newblock \textit{arXiv preprint arXiv:1009.3857}.

\bibitem[{Seguy et~al.(2017)Seguy, Damodaran, Flamary, Courty, Rolet and
  Blondel}]{seguy2017large}
\textsc{Seguy, V.}, \textsc{Damodaran, B.~B.}, \textsc{Flamary, R.},
  \textsc{Courty, N.}, \textsc{Rolet, A.} and \textsc{Blondel, M.} (2017).
\newblock Large-scale optimal transport and mapping estimation.
\newblock \textit{arXiv preprint arXiv:1711.02283}.

\bibitem[{Seguy et~al.(2018)Seguy, Damodaran, Flamary, Courty, Rolet and
  Blondel}]{seguy2018large}
\textsc{Seguy, V.}, \textsc{Damodaran, B.~B.}, \textsc{Flamary, R.},
  \textsc{Courty, N.}, \textsc{Rolet, A.} and \textsc{Blondel, M.} (2018).
\newblock Large scale optimal transport and mapping estimation.
\newblock In \textit{International Conference on Learning Representations}.
\newline\urlprefix\url{https://openreview.net/forum?id=B1zlp1bRW}

\bibitem[{Sobel(1990)}]{sobel1990isotropic}
\textsc{Sobel, I.} (1990).
\newblock An isotropic 3$\times$ 3 image gradient operater.
\newblock \textit{Machine vision for three-dimensional scenes} 376--379.

\bibitem[{Solomon et~al.(2015)Solomon, De~Goes, Peyr{\'e}, Cuturi, Butscher,
  Nguyen, Du and Guibas}]{solomon2015convolutional}
\textsc{Solomon, J.}, \textsc{De~Goes, F.}, \textsc{Peyr{\'e}, G.},
  \textsc{Cuturi, M.}, \textsc{Butscher, A.}, \textsc{Nguyen, A.}, \textsc{Du,
  T.} and \textsc{Guibas, L.} (2015).
\newblock Convolutional wasserstein distances: Efficient optimal transportation
  on geometric domains.
\newblock \textit{ACM Transactions on Graphics (TOG)}, \textbf{34} 66.

\bibitem[{Villani(2008)}]{villani2008optimal}
\textsc{Villani, C.} (2008).
\newblock \textit{Optimal transport: old and new}, vol. 338.
\newblock Springer Science \& Business Media.

\bibitem[{Xie et~al.(2018)Xie, Wang, Wang and Zha}]{xie2018fast}
\textsc{Xie, Y.}, \textsc{Wang, X.}, \textsc{Wang, R.} and \textsc{Zha, H.}
  (2018).
\newblock A fast proximal point method for wasserstein distance.
\newblock \textit{arXiv preprint arXiv:1802.04307}.

\bibitem[{Yang and Uhler(2018)}]{yang2018scalable}
\textsc{Yang, K.~D.} and \textsc{Uhler, C.} (2018).
\newblock Scalable unbalanced optimal transport using generative adversarial
  networks.
\newblock \textit{arXiv preprint arXiv:1810.11447}.

\bibitem[{Zhang et~al.(2016)Zhang, Bengio, Hardt, Recht and
  Vinyals}]{zhang2016understanding}
\textsc{Zhang, C.}, \textsc{Bengio, S.}, \textsc{Hardt, M.}, \textsc{Recht, B.}
  and \textsc{Vinyals, O.} (2016).
\newblock Understanding deep learning requires rethinking generalization.
\newblock \textit{arXiv preprint arXiv:1611.03530}.

\bibitem[{Zhao et~al.(2016)Zhao, Mathieu and LeCun}]{zhao2016energy}
\textsc{Zhao, J.}, \textsc{Mathieu, M.} and \textsc{LeCun, Y.} (2016).
\newblock Energy-based generative adversarial network.
\newblock \textit{arXiv preprint arXiv:1609.03126}.

\bibitem[{Zhu et~al.(2017)Zhu, Park, Isola and Efros}]{zhu2017unpaired}
\textsc{Zhu, J.-Y.}, \textsc{Park, T.}, \textsc{Isola, P.} and \textsc{Efros,
  A.~A.} (2017).
\newblock Unpaired image-to-image translation using cycle-consistent
  adversarial networks.
\newblock \textit{arXiv preprint}.

\end{thebibliography}
\bibliographystyle{ims}


\newpage
\onecolumn
\appendix
\textbf{\Large{Appendix}}

\section{Network Architecture}
\label{sec:app_network}

\subsection{No-sharing Network}
\label{sec:no_share_structure}
The CNN architecture for experiments in Section \ref{sec:section53}. Table~\ref{tab:CNN_gen_noshare} shows the architecture of two mappings $G_X$ and $G_Y$. The two mappings have identical architechture. 
\begin{table}[ht!]
	\caption{The CNN architecture for experiments of real datasets in Section \ref{sec:section53}.}\label{tab:CNN_gen_noshare}
	\begin{center}
			\begin{tabular}{ l l@{ }l@{ }l@{ }l l} 
				\hline
				\hline
				\textbf{Input}:&\multicolumn{4}{l}{${z}$ $\in$ $\mathbb{R}^{100}$ $\sim$ $\mathcal{N}$(0,  $\textit{I}$ )}&\\ 
				\hline
				& \multicolumn{4}{l}{Convolution Filter} & Activation \\
				\hline
				Deconv:& [4 $\times$ 4, &512, &stride = 1, &padding=0] &BN, ReLU \\
				\hline
				Deconv:&[4 $\times$ 4, &256, &stride = 2, &padding=1]   &BN, ReLU \\
				\hline
				Deconv:&[4 $\times$ 4, &128, &stride = 2, &padding=1]   &BN, ReLU \\
				\hline
				Deconv:&[4 $\times$ 4, &64, &stride = 2, &padding=1]   &BN, ReLU \\
				\hline
				Deconv:&[4 $\times$ 4, &3, &stride = 2, &padding=1]   &Tanh \\
				\hline
				\hline
		\end{tabular}
	\end{center}
\end{table}

Table~\ref{tab:CNN_dis1} shows the architecture of two discriminators $\lambda_X,\lambda_Y$. The two networks have identical architechture and do not share parameters. 
\begin{table}[ht!]
	\caption{The CNN architecture of $\lambda_X,\lambda_Y$ for experiments of real datasets in Section \ref{sec:section53}. }\label{tab:CNN_dis1}
	\begin{center}
		\begin{tabular}{ l l@{ }l@{ }l@{ }l l } 
			\hline
			\hline
			\textbf{Input}:&\multicolumn{4}{l}{Image ${x}$ $\in$ $\mathbb{R}^{64 \times 64 \times 3}$ $\sim \mu$ or $\nu$}& \\ 
			\hline
			& \multicolumn{4}{l}{Convolution Filter} & Activation \\
			\hline
			Conv:& [4 $\times$ 4, &64, &stride = 1, &padding=0] &ReLU\\
			\hline
			Conv:& [4 $\times$ 4, &128, &stride = 2, &padding=1] &BN, ReLU\\
			\hline
			Conv:& [4 $\times$ 4, &256, &stride = 2, &padding=1] &BN, ReLU \\
			\hline
			Conv:& [4 $\times$ 4, &512, &stride = 2, &padding=1] &BN, ReLU \\
			\hline
			Conv:& [4 $\times$ 4, &1, &stride = 1, &padding=0] &$-$ \\
			\hline
			\hline
		\end{tabular}
	\end{center}
\end{table}

\subsection{Convolutional Network}
\label{sec:cnn_structure}
The CNN architecture for USPS, MNIST and MNISTM. PReLU activation is applied \cite{he2015delving}. Table~\ref{tab:CNN_gen} shows the architecture of two generators $G_X$ and $G_Y$. The last column in Table~\ref{tab:CNN_gen} means whether  $G_X$ and $G_Y$ share the same parameter.

\begin{table}[ht!]
	\caption{The CNN generater architecture for USPS, MNIST and MNISTM. $ch=1$ for USPS and MNIST; $ch=3$ for MNISTM. }\label{tab:CNN_gen}
	\begin{center}
			\begin{tabular}{ l l@{ }l@{ }l@{ }l l l} 
				\hline
				\hline
				\textbf{Input}:&\multicolumn{4}{l}{${z}$ $\in$ $\mathbb{R}^{100}$ $\sim$ $\mathcal{N}$(0,  $\textit{I}$ )}& &\\ 
				\hline
				& \multicolumn{4}{l}{Convolution Filter} & Activation & Shared\\
				\hline
				Deconv:& [4 $\times$ 4, &1024, &stride = 1, &padding=0] &BN, PReLU & True\\
				\hline
				Deconv:&[3 $\times$ 3, &512, &stride = 2, &padding=1]   &BN, PReLU & True\\
				\hline
				Deconv:&[3 $\times$ 3, &256, &stride = 2, &padding=1]   &BN, PReLU & True\\
				\hline
				Deconv:&[3 $\times$ 3, &128, &stride = 2, &padding=1]   &BN, PReLU & True\\
				\hline
				Deconv:&[3 $\times$ 6, &ch, &stride = 1, &padding=1]   &Sigmoid & False\\
				\hline
				\hline
		\end{tabular}
	\end{center}
\end{table}

Table~\ref{tab:CNN_dis} shows the architecture of two discriminators $\lambda_X,\lambda_Y$, and two classifiers $D_X$, $D_Y$. The last column in Table~\ref{tab:CNN_gen} uses $(\cdot,\cdot)$ to denote which group of discriminators share the same parameter.
\begin{table}[ht!]
	\caption{The CNN discriminator architecture for USPS, MNIST and MNISTM. $ch=1$ for USPS and MNIST; $ch=3$ for MNISTM. $ch_o=1$ for $\lambda_X$ and $\lambda_Y$; $ch_o=10$ for $D_X$ and $D_Y$. }\label{tab:CNN_dis}
	\begin{center}
		\begin{tabular}{ l l@{ }l@{ }l@{ }l l l} 
			\hline
			\hline
			\textbf{Input}:&\multicolumn{4}{l}{Image ${x}$ $\in$ $\mathbb{R}^{28 \times 28 \times ch}$ $\sim \mu$ or $\nu$}& &\\ 
			\hline
			& \multicolumn{4}{l}{Convolution Filter} & Activation & Shared\\
			\hline
			Conv:& [5 $\times$ 5, &20, &stride = 1, &padding=0] &MaxPooling(2,2) & $(\lambda_{X},D_X)$;$(\lambda_{Y},D_Y)$\\
			\hline
			Conv:& [5 $\times$ 5, &50, &stride = 1, &padding=0] &MaxPooling(2,2) & $(\lambda_{X},\lambda_{Y},D_X,D_Y)$\\
			\hline
			Conv:& [4 $\times$ 4, &500, &stride = 1, &padding=0] &PReLU & $(\lambda_{X},\lambda_{Y},D_X,D_Y)$\\
			\hline
			Conv:& [1 $\times$ 1, &$ch_o$, &stride = 1, &padding=0] &$-$ & $(\lambda_{X});(\lambda_{Y});(D_X,D_Y)$\\
			\hline
			\hline
		\end{tabular}
	\end{center}
\end{table}

\subsection{Residual Network}
\label{sec:resnet_structure}

The ResNet architecture for SVHN $\rightarrow$ MNIST. Table~\ref{tab:resnet_gen} shows the architecture of two generators $G_X$ and $G_Y$. The last column in Table~\ref{tab:resnet_gen} means whether  $G_X$ and $G_Y$ share the same parameter. The Residual block is the same as the one in \citet{miyato2018spectral}.

\begin{table}[ht!]
	\caption{The ResNet generater architecture for  SVHN $\rightarrow$ MNIST. $ch=1$ for MNIST; $ch=3$ for SVHN. }\label{tab:resnet_gen}
	\begin{center}
		\begin{tabular}{ l@{ }l ll} 
			\hline
			\hline
			\textbf{Input}:&${z}$ $\in$ $\mathbb{R}^{100}$ $\sim$ $\mathcal{N}$(0,  $\textit{I}$ )&&\\ 
			\hline 
			&Layer Size& Activation& Shared\\
			\hline
			Linear:& 100 $\rightarrow $ $4$ $\times$ $4$$\times$ 128 &$-$&True\\
			\hline
			ResBlocks:& [128, Up-sampling] &$-$&True\\
			\hline
			ResBlocks:& [128, Up-sampling] &$-$&True\\
			\hline
			ResBlocks:& [128, Up-sampling] &BN,PReLU&True\\
			\hline
			Conv:& [3 $\times$ 3, $ch$, stride = 1, padding =0]& Sigmoid&False \\
			\hline
			\hline
		\end{tabular}
	\end{center}
\end{table}

Table~\ref{tab:resnet_dis} shows the architecture of two discriminators $\lambda_X,\lambda_Y$, and two classifiers $D_X$, $D_Y$. The last column in Table~\ref{tab:resnet_dis} uses $(\cdot,\cdot)$ to denote which group of discriminators share the same parameter.

\begin{table}[ht!]
	\caption{The ResNet discriminator architecture for  SVHN $\rightarrow$ MNIST. $ch=1$ for MNIST; $ch=3$ for SVHN.  $ch_o=1$ for $\lambda_X$ and $\lambda_Y$; $ch_o=10$ for $D_X$ and $D_Y$. }\label{tab:resnet_dis}
	\begin{center}
		\begin{tabular}{ l@{ }l ll} 
			\hline
			\hline
			\textbf{Input}:&Image ${x}$ $\in$ $\mathbb{R}^{28 \times 28 \times ch}$ $\sim \mu$ or $\nu$&&\\ 
			\hline 
			&Layer Size& Activation& Shared\\
			\hline
			ResBlocks:& [128, Down-Sampling] &$-$&$(\lambda_{X},D_X)$;$(\lambda_{Y},D_Y)$\\
			\hline
			ResBlocks:& [128, Down-Sampling] &$-$&$(\lambda_{X},\lambda_{Y},D_X,D_Y)$\\
			\hline
			ResBlocks:& [128, Down-Sampling] &$-$&$(\lambda_{X},\lambda_{Y},D_X,D_Y)$\\
			\hline
			Conv:& [4 $\times$ 4, 500, stride = 1, padding=0] &PReLU & $(\lambda_{X},\lambda_{Y},D_X,D_Y)$\\
			\hline
			Conv:& [1 $\times$ 1, $ch_o$, stride = 1, padding=0] &$-$ & $(\lambda_{X});(\lambda_{Y});(D_X,D_Y)$\\
			\hline
			\hline
		\end{tabular}
	\end{center}
\end{table}

\end{document}